\documentclass{article}

\usepackage{multirow}

\usepackage{iclr2023_conference, times}

\usepackage[utf8]{inputenc} 
\usepackage[T1]{fontenc}    
\usepackage{hyperref}       
\usepackage{url}            
\usepackage{booktabs}       
\usepackage{amsfonts}       
\usepackage{nicefrac}       
\usepackage{microtype}      
\usepackage{xcolor}   
\usepackage{graphicx}
\usepackage{makecell}
\usepackage{comment}
\usepackage{amsmath}

\usepackage{graphicx}
\usepackage{subfig}

\newcommand{\mat}[1]{\mathbf{#1}}

\def\ie{{\frenchspacing\it i.e.}}

\title{Seeing is Believing: Brain-inspired Modular Training for Mechanistic Interpretability}

\author{Ziming Liu, Eric Gan \& Max Tegmark \\
Department of Physics, Institute for AI and Fundamental Interactions, MIT \\
\texttt{\{zmliu,ejgan,tegmark\}@mit.edu}
}

\begin{document}

\maketitle

\begin{abstract}
    We introduce Brain-Inspired Modular Training (BIMT), a  method for making neural networks more modular and interpretable. Inspired by brains, BIMT embeds neurons in a geometric space and augments the loss function with a cost proportional to the length of each neuron connection. We demonstrate that BIMT discovers useful modular neural networks for many simple tasks, 
    revealing compositional structures in symbolic formulas, interpretable decision boundaries and features for classification, and mathematical structure in algorithmic datasets. 
    The ability to directly \textit{see} modules with the naked eye can complement current mechanistic interpretability strategies such as probes, interventions or staring at all weights. 
\end{abstract}

\section{Introduction}

Although deep neural networks have achieved great successes, mechanistically interpreting them remains quite challenging~\citep{olah2020zoom, olsson2022context, michaud2023quantization, elhage2021mathematical, wang2023interpretability}. If a neural network can be decomposed into smaller modules~\citep{olah2020zoom}, interpretability may become much easier.

In contrast to artificial neural networks, brains are remarkably modular~\citep{bear2020neuroscience}. 
We conjecture that this is because artificial neural networks (e.g., fully connected neural networks) have a symmetry that 
brains lack: both the loss function and the most popular regularizers are invariant under permutations of neurons in each layer.
In contrast, the cost of connecting two biological neurons depends on how far apart they are, because an axon needs to traverse this distance, thereby using energy and brain volume and causing time delay. 



\begin{figure}[tbp]
    \centering
    \includegraphics[width=1\linewidth]{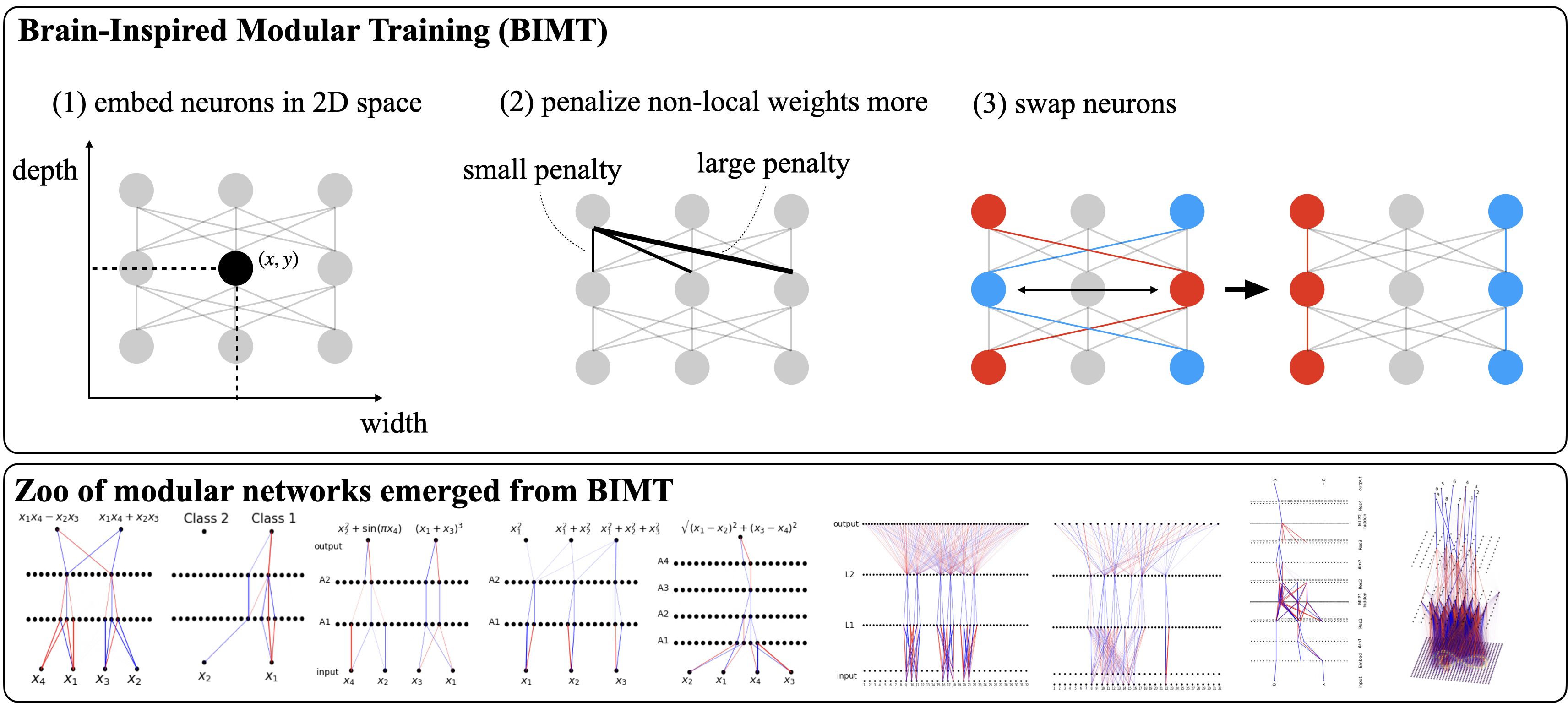}
    \caption{Top: Brain-inspired modular training (BIMT) contains three ingredients: (1) embedding neurons into a geometric space (e.g., 2D Euclidean space); (2) training with regularization which penalizes non-local weights more; (3) swapping neurons during training to further enhance locality. Bottom: Zoo of modular networks obtained via BIMT (see experiments for details).}
    \label{fig:BIMT}
\end{figure}

To facilitate the discovery of more modular and interpretable neural networks, we introduce Brain-Inspired Modular Training (BIMT). Inspired by brains, we embed neurons in a geometric space where distances are defined, and augment the loss function with a cost proportional to the length of each neuron connection times the absolute value of the connection weight. This obviously encourages {\it locality}, \ie, keeping neurons that need to communicate as close together as possible. Any Riemannian manifold can be used; we explore 2D and 3D Euclidean space for easy visualization (see Figure~\ref{fig:BIMT}).

We demonstrate the power of BIMT on a broad range of tasks, finding that it can reveal interesting and sometimes unexpected structures. On symbolic formula datasets, BIMT is able to discover  structures such as independence, compositionality and features sharing, which are useful for scientific applications. For classifications tasks, we find that BIMT may produce interpretable decision boundaries and features. For algorithmic tasks, we find BIMT to produce tree-like connectivity graphs, not only supporting the group representation argument in~\cite{chughtai2023toy}, but also revealing a (somewhat unexpected) mechanism where multiple modules vote. Although most of our experiments are conducted on fully connected networks for vector inputs, we also conduct experiments demonstrating that BIMT generalizes to other types of data
(e.g., images) and architectures (e.g., transformers). 

This paper is organized as follows: Section \ref{sec:method} introduces brain-inspired modular training (BIMT). Section \ref{sec:experiments} applies BIMT to various tasks, demonstrating its interpretability power. We describe related work in Section \ref{sec:related_works} and discuss our conclusions in Section \ref{sec:conclusions}.

\section{Brain-Inspired Modular Training (BIMT)}\label{sec:method}

Human brains are modular and sparse, which is arguably the reason why they are so efficient. To make neural networks more efficient, it is desirable to make them modular and sparse, just like our brains. Sparsity is a well-studied topic in neural networks, and can be encouraged by including $L_1/L_2$ penalty in training or by applying pruning to model weights~\citep{han2015learning,anwar2017structured}. As for modularity, most of research explicitly introduce modules~\citep{pfeiffer2023modular, kirsch2018modular}, but this requires prior knowledge about problem structures. Our motivation question is thus:

\begin{center}
    \fbox{Q: What training techniques can induce modularity in otherwise non-modular networks?}
\end{center}

In other words, our goal is to let modularity emerge from non-modular networks when possible. In this section, we propose a method called Brain-Inspired Modular Training (BIMT), which explicitly steers neural networks to become more modular and sparse during training. BIMT consists of three key ingredients (see Figure~\ref{fig:BIMT}): (1) embedding the network to a geometric space; (2) training to encourage locality and sparsity; (3) swapping neurons for better locality.

{\bf Notation} For simplicity we describe how to do BIMT with fully connected networks; generalization to other architectures is possible. We distinguish between \textit{weight layers} and \textit{neuron layers}. Assuming a fully connected network to have $L$ weight layers, whose $i^{\rm th}$ weight layer $(i=1,\cdots,L)$ has weights $\mat{W}_i\in\mathbb{R}^{n_{i-1}\times n_i}$ and biases $\mat{b}_i\in\mathbb{R}^{n_i}$, where $n_{i-1}$ and $n_i$ are the number of neurons incoming to and outgoing from the $i^{\rm th}$ weight layer. The $i^{\rm th}$\ $(i=0,\cdots,L)$ neuron layer has $n_i$ neurons. The input and output dimension of the whole network is $n_0$ and $n_L$, respectively. 

{\bf Step 1: Embedding the network to a geometric space} We now embed the whole network into a space where the $j^{\rm th}$ neuron in the $i^{\rm th}$ layer is the $(i,j)$ neuron located at $\mat{r}_{ij}$. If this is 2D Euclidean space, neurons in the same neuron layer share the same $y$-coordinate and are uniformly spaced in $x\in[0,A] (A>0)$. Different neuron layers are vertically separated by a distance $y_*>0$, so
\begin{equation}
\mat{r}_{ij}\equiv (x_{ij},y_{ij})=(Aj/n_i, iy_*).
\end{equation}
The weight that connects the $(i-1,j)$ neuron and the $(i,k)$ neuron has value $w_{ijk}\equiv (\mat{W}_i)_{jk}$, and the bias at the $(i+1,k)$ neuron is $b_{ik}\equiv (\mat{b}_i)_k$ and its length is defined as
\begin{equation}
    d_{ijk} \equiv \left|\mat{r}_{i-1,j}-\mat{r}_{ik}\right|.
\end{equation}
We will use $L_1$-norm, giving $d_{ijk}=A|x_{i-1,j}-x_{ik}|+y_*$, but other vector norms can also be used. For example, $L_2$-norm gives $d_{ijk}=\left(A^2|x_{i-1,j}-x_{ik}|^2+y_*^2\right)^{1/2}$.

\begin{figure}[tbp]
\includegraphics[width=1.0\linewidth]{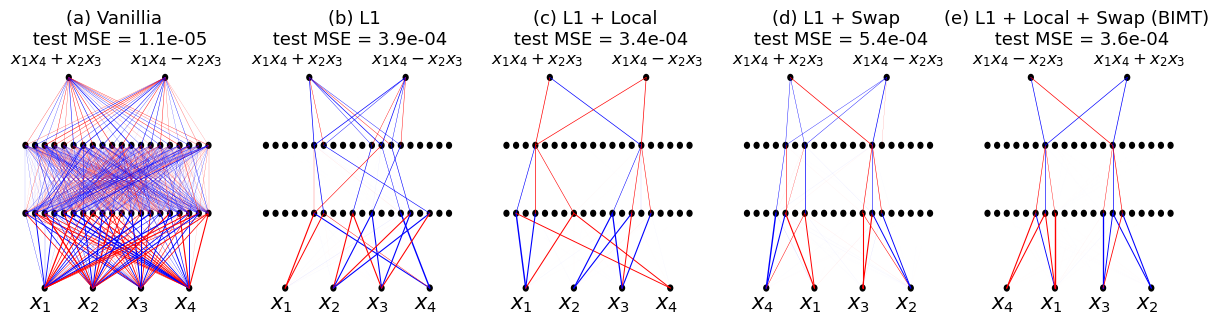}
\caption{The connectivity graphs of neural networks when trained with different techniques for a regression problem (blue/red denote positive/negative weights). Our proposed BIMT = $L_1$ regularization (not novel) + local regularization (novel) + swap (novel). BIMT finds the simplest circuit (e) which clearly contains two parallel modules, with a small sacrifice in test loss compared to vanilla (a), but with lower loss than for mere $L_1$ regularization (b). Note that swapping aims to reduce the local connection cost, so all of (c)(d)(e) encourage locality.}
\label{fig:sf_method}
\end{figure}

{\bf Step 2: imposing regularization that encourage locality} We define the connection cost for weight and bias parameters of the whole network to be
\begin{equation}
    \ell^w = \sum_{i=1}^{L}\sum_{j=1}^{n_{i}}\sum_{k=1}^{n_{i+1}} d_{ijk}|w_{ijk}|,\quad  \ell^b = \sum_{i=1}^L\sum_{j=1}^{n_i} y_* |b_{ij}|.
\end{equation}
When training for a particular task, in addition to the prediction loss $\ell^{\rm pred}$, we include $\ell^w$ and $\ell^b$ as regularizations:
\begin{equation}
    \ell = \ell^{\rm pred} + \lambda(\ell^w + \ell^b),
\end{equation}
where $\lambda$ is the strength of the regularization. Without loss of generality, we can set $y_*=1$, leaving only two hyper-parameters $\lambda$ and $A$. Setting $A=0$ reduces to standard $L_1$ regularization which solely encourages sparsity. $A>0$ further encourages locality, in addition to sparsity. 

{\bf Step 3: swapping neurons for better locality} We encourage locality (reduction of $\ell^w$) not only by updating weights via gradient descent, but also by swapping two neurons in the same neuron layer (i.e., swapping corresponding incoming/outgoing weights), when this reduces $\ell^w$. Gradient descent (continuous search) can get stuck at bad local minima where non-local connections are still present (see Figure~\ref{fig:sf_method} (c)), while swapping (discrete search) can avoid this (see Figure~\ref{fig:sf_method} (e)). Such swapping leaves the function implemented by the whole network (hence $\ell^{\rm pred}$) unchanged, but improves locality (see Figure~\ref{fig:BIMT} right). However, trying every possible permutation is prohibitively expensive. We assign each neuron $(i,j)$ a score $s_{ij}$ to indicate its importance:
\begin{equation}
    s_{ij} = \sum_{p=1}^{n_{i-1}} |w_{ipj}| + \sum_{q=1}^{n_{i+1}} |w_{i+1,jq}|,
\end{equation}
which is the sum of (absolute values) of incoming and outgoing weights. We sort neurons in the same layer based on their scores and define neurons the top $k$-scores as "important" neurons. For each important neuron, we swap it with the neuron in the same layer causing the greatest decrease in $l^w$ if it helps. Since swaps are somewhat expensive, requiring $O(nkL)$ computations, we implement swaps only every $S\gg 1$ training steps. We allow swaps also of input and output neurons, if not stated otherwise.

{\bf BIMT = ${\boldsymbol L_1}$ + Local + Swap} To summarize, BIMT means \underline{local} \underline{$L_1$ regularization} with \underline{swaps}. Both "local" and "swap" are novel contributions of this paper, while $L_1$ regularization is quite standard. If one wants to ablate "local" or "swap", one can set $A=0$ to remove "local", or set $S\to \infty$ to remove "swap". Our experience is that the joint use of "local" and "swap" usually gives the most interpretable networks. As a simple case, we compare BIMT to baselines on a regression problem, shown in Figure~\ref{fig:sf_method}. On top of $L_1$, although using "local" or "swap" alone gives reasonably interpretable networks, the joint use of both produces the most interpretable network (at least visually). Although using $L_1$ alone leads to a reasonably sparse network, the network is neither modular nor optimally sparse (see Appendix \ref{app:pruning} for pruning results).

{\bf Connectivity graphs} As in Figure \ref{fig:sf_method} and throughout the paper, we will use connectivity graphs to visualize neural network structures. For visualization purposes, we normalize weights by the max absolute value in the same layer (so the normalized values lie in range $[-1,1]$). A weight is displayed as a line connecting two neurons, with its thickness proportional to its normalized value, with its color to be blue (red) if the value is positive (negative). Note that  we draw all weights and do not explicitly ignore small weights. The reason why connectivity graphs appear sparse is because naked eyes cannot identify too thin lines.

\section{Experiments}\label{sec:experiments}

In this section, we apply BIMT to a wide range of applications. In all cases, BIMT can result in modular and sparse networks, which immediately provide interpretability on the microscopic level and the macroscopic level. At the microscopic level, we can understand which neurons are useful, what each useful neuron is doing,  where/how information of interest is located/computed. At the macroscopic level, we can understand relations between different modules (e.g., in succession or in parallel), and how they cooperate to make the final prediction. From Section 3.1 to 3.3, we train fully-connceted neural networks with BIMT for regression, classification and algorithmic tasks. In Section 3.4, we show that BIMT can generalize to transformers and demonstrate it in in-context learning. In Section 3.5, we demonstrate that BIMT can easily go beyond vector-type data to tensor-type data (e.g., images). In general, BIMT achieves interpretability with no or modest drop in performance, summarized in Table~\ref{tab:tradeoff}. All experiments are runnable on a cpu (M1) usually in minutes (at most 2 hours).

\begin{table}[htbp]
    \centering
    \caption{BIMT achieves interpretability with no or modest performance drop}
    \resizebox{\columnwidth}{!}{%
    \begin{tabular}{ccccccccc}\hline
    dataset  &  \makecell{symbolic \\ (a)} & \makecell{symbolic \\ (b)} & \makecell{symbolic \\ (c)} & \makecell{two \\ moon} & \makecell{modular \\ addition} & permutation & \makecell{in-context} & MNIST \\\hline
    metric   & loss & loss & loss & accuracy & accuracy & accuracy & loss & accuracy  \\\hline
    without BIMT & 5.8e-3 & 1.1e-5 & 1.2e-4 & 100.0\% & 100.0\% & 100.0\% & 7.2e-5 & 98.5\% \\
    with BIMT & 7.4e-3 & 8.5e-5 & 1.3e-3 & 100.0\% & 100.0\% & 100.0\% & 1.8e-4 & 98.0\% \\\hline
    \end{tabular}
    }
    \label{tab:tradeoff}
\end{table}

\subsection{Symbolic Formulas}

\begin{figure}[htbp]
\includegraphics[width=1.0\linewidth]{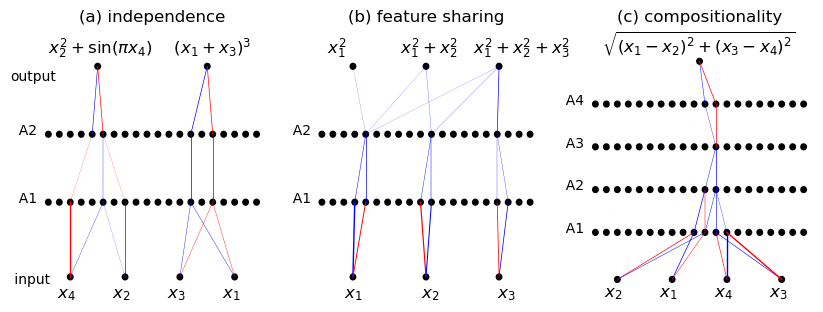}
\caption{The connectivity graphs of neural networks trained with BIMT to regress symbolic formulas (blue/red lines stand for positive/negative weights). For symbolic formulas with modular properties, e.g., independence, shared features or compositionality, the connectivity graphs display modular structures revealing these properties.}
\label{fig:sf_various}
\end{figure}

Symbolic formulas are prevalent in scientific domains. In recent years, as increasingly more data are collected from experiments, it is desirable to distill symbolic formulas from experimental data, a task called symbolic regression~\citep{udrescu2020ai,udrescu2020ai2,cranmer2020discovering}. However, symbolic regression usually faces an all-or-nothing situation, i.e., either succeeds gloriously or fails miserably. Consequently, a tool supplementary to symbolic regression is called for, which can robustly reveal the high-level structure of formulas. We show below that BIMT can discover such structures in formulas.

We consider the task of predicting $\mat{y}=(y_1,\cdots,y_{d_o})$ from $\mat{x}=(x_1,\cdots,x_{d_i})$ where $y_i=f_i(\mat{x})$ are symbolic functions. We randomly sample each $\mat{x}_i$ from $U[-1,1]$ and compute $y_i=f_i(\mat{x})$ to generate the dataset. We use fully-connected networks with SiLU activations (architectures shown in Figure~\ref{fig:sf_various}), training networks using the MES loss with the Adam optimizer with learning rate $10^{-3}$ for $20000$ steps, while choosing $A=2$, $y_*=0.1$, $k=6$, and $S=200$. We schedule $\lambda$ as such: $(10^{-3},10^{-2},10^{-3})$ for (5000, 10000, 5000) steps.

We apply BIMT to several formulas, each of which has certain modular properties, as shown in Figure~\ref{fig:sf_various}. (a) {\bf independence}.  $y_1=x_2^2+{\rm sin}(\pi x_4)$ is independent of $x_1$ and $x_3$, while $y_2=(x_1+x_3)^3$ is independent of $x_2$ and $x_4$. As desired, BIMT results in a network splitted into two parallel modules independent of each other, one only involving $(x_1,x_3)$, the other only involving $(x_2,x_4)$. (b) {\bf feature sharing}. For targets $(y_1,y_2,y_3)=(x_1^2,x_1^2+x_2^2,x_1^2+x_2^2+x_3^2)$, learning shared features $(x_1^2,x_2^2,x_3^2)$ is beneficial for predicting all targets. Indeed, in the neuron layer A2, the only three active neurons correspond to these shared features (see Appendix \ref{app:sf}). (c) {\bf compositionality}. Computing $y=\sqrt{(x_1-x_2)^2+(x_3-x_4)^2}\equiv \sqrt{I}$ requires computing $I$ first, which is an important intermediate quantity. We find that the only active neuron in layer A3 has activations highly correlated with $I$ . Although one might worry that these extremely sparse networks could severely underfit, we show that fitting is reasonably good in Appendix \ref{app:sf}.

\subsection{Two moon classification}\label{sec:exp_two_moon}

\begin{figure}[htbp]
    \centering
    \includegraphics[width=0.9\linewidth]{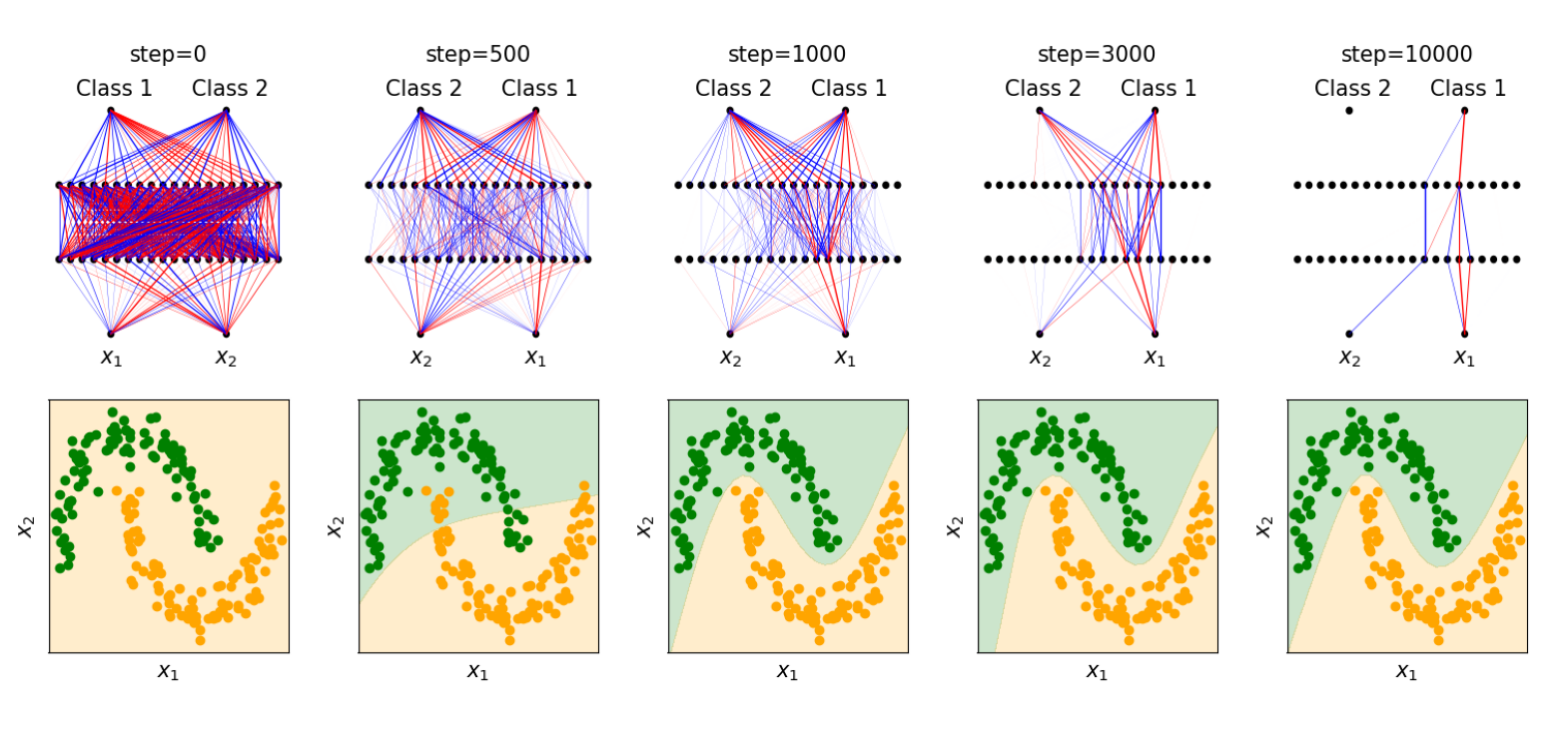}
    \caption{Top: Evolution of network structures trained with BIMT on the two moon dataset. Bottom: Evolution of decision boundaries. }
    \label{fig:two_moon_evolution}
\end{figure}

Interpretable decision boundaries help to make classification trustworthy. Moreover, decision boundaries with fewer pieces are more likely to generalize better. So it is desirable that neural networks used for classifications are sparse and interpretable. 

We apply BIMT to the toy two moon dataset~\citep{scikit-learn}. The architecture is shown in Figure~\ref{fig:two_moon_evolution} (the final softmax layer is not shown), with the same training details used in Section 3.1, the only difference being using cross entropy loss. The evolution of the neural network is shown in Figure~\ref{fig:two_moon_evolution}: Starting from a (randomly initialized) dense network, the network becomes increasingly sparse and modular, ending up as a network with only 6 useful hidden neurons. We can roughly split the training process into three phases: (i) in the first phase (step 0 to 1000), the neural network mainly aims to fit the data while slightly sparsifying the network; (ii) in the second phase (step 1000 to 3000), the neural network sparsifies the network in a symmetric way (both outputs of class 1 and 2 have neurons connecting to them). (iii) in the third phase (step 3000 to end), the network prunes itself to become asymmetric, with useful neurons only connecting to Class 1 output. In Appendix \ref{app:two_moon}, we interpret what each weight is doing by editing them (zeroing) and see how this affects decision boundaries. 

\subsection{Algorithmic datasets}
\begin{figure}
    \centering
    \includegraphics[width=0.5\linewidth]{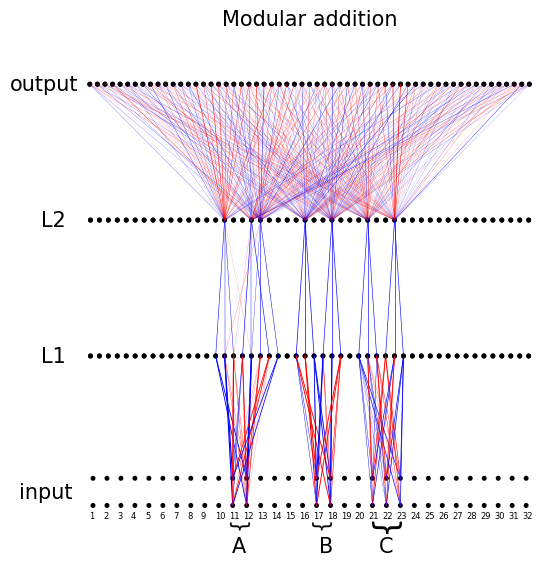}
    \includegraphics[width=0.15\linewidth]{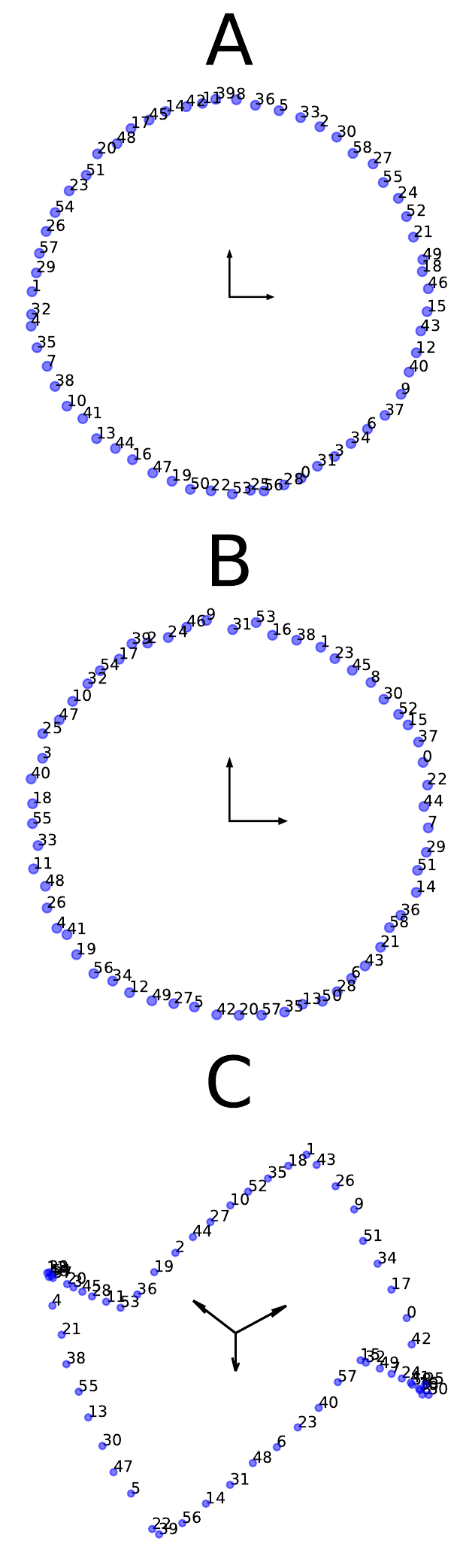}\quad
    \begin{tabular}[b]{|c|c|}\hline
      Knockout & Accuracy \\ \hline
      None & 100.00\% \\\hline
      A & 15.25\% \\
      B & 29.33\% \\
      C & 33.67\% \\
      A, B & 3.39\% \\
      A, C & 5.08\% \\
      B, C & 10.28\% \\
      A, B, C & 1.69\% \\\hline
      A11 & 47.11\% \\
      A12 & 46.51\% \\
      B17 & 50.47\% \\
      B18 & 51.42\% \\
      C21 & 77.10\% \\
      C22 & 73.60\% \\
      C23 & 78.17\%  \\\hline
      \makecell{All but\\ A, B, C} & 100.00\% \\
      \hline
    \end{tabular}
    \caption{MLP trained with BIMT for modular addition. Left: the final connectivity graph is tree-like, demonstrating three parallel modules (voters); middle: the representations of each module in the input layer; right: ablation results, which imply a voting mechanism. The input layer contains embeddings of two tokens, which overlap each other but are drawn to be vertically separated.}
    \label{fig:modadd}
\end{figure}

Algorithmic datasets are ideal for benchmarking mechanistic interpretability methods, because they are mathematically well understood. Consider a binary operation $a\circ b=c$ ($a,b,c$ are discrete and treated as tokens) and a neural network is tasked with predicting $c$ from embeddings of $a$ and $b$. For modular addition, ~\cite{liu2022towards} discovers that ring-like representations emerge in training. ~\cite{nanda2023progress} reverse engineered these networks, finding that the network internally implements trigonometric identities. For more general group operations,~\cite{chughtai2023toy} suggests that representation theory is key for neural networks to generalize. However, in these papers it is usually not obvious which neurons are useful, or what the overall modular structure of the network is. Since BIMT explicitly optimizes modularity, it is able to produce networks which self-reveal their structure.

{\bf Modular addition} The task is to predict $c$ from $(a,b)$, where $a+b=c\ ({\rm mod}\ 59)$. Each token $a$ is embedded as a $d=32$-dimensional vector $\mat{E}_a$, initialized as a random normal vector at initialization and trainable later. The concatenation of $\mat{E}_a$ and $\mat{E}_b$ is fed to a two-hidden-layer MLP, shown in Figure~\ref{fig:modadd}. We split train/test 80\%/20\%. We train the network with BIMT with cross entropy loss using the Adam optimimizer (lr $=10^{-3}$) for 20,000 steps. We choose $A=2$, $y_*=0.5$, $k=30$, and $S=200$. We schedule $\lambda$ as such: $(0.1,1,0.1)$ for (5000, 10000, 5000) steps. 

After training, the network looks like a tree with three roots (A, B, C), shown in Figure~\ref{fig:modadd}. We visualize embeddings corresponding to these roots (modules), finding that the token embeddings form circles in 2D (A, B) and a bow tie in 3D (C). In contrast to~\cite{liu2022towards} and ~\cite{nanda2023progress} where post-processing (e.g., principal component analysis) is needed to obtain ring-like representations, the ring structures here automatically align to privileged bases, which is probably because embeddings are also regularized with $L_1$. To evaluate how these parallel modules are important for making predictions, we compute accuracy after knocking out some of them. The result is quite surprising: knocking out one of the modules can severely degrade the performance (from 100\% to 15.25\%, 29.33\% and 33.67\% for knocking out A, B or C). This means that modules are cooperating together to make predictions correct, similar to majority voting for error correction. To verify the universality of this argument, we include more tree graphs for perturbed initializations and different random seeds in Appendix~\ref{app:modadd_sensitivity}.

\begin{figure}
    \centering
    \includegraphics[width=0.5\linewidth]{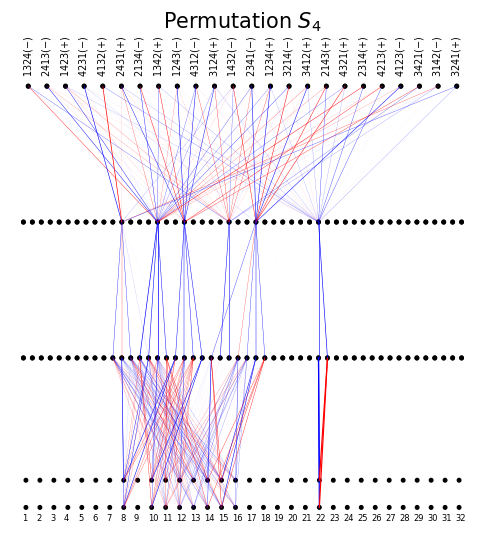}
    \includegraphics[width=0.48\linewidth]{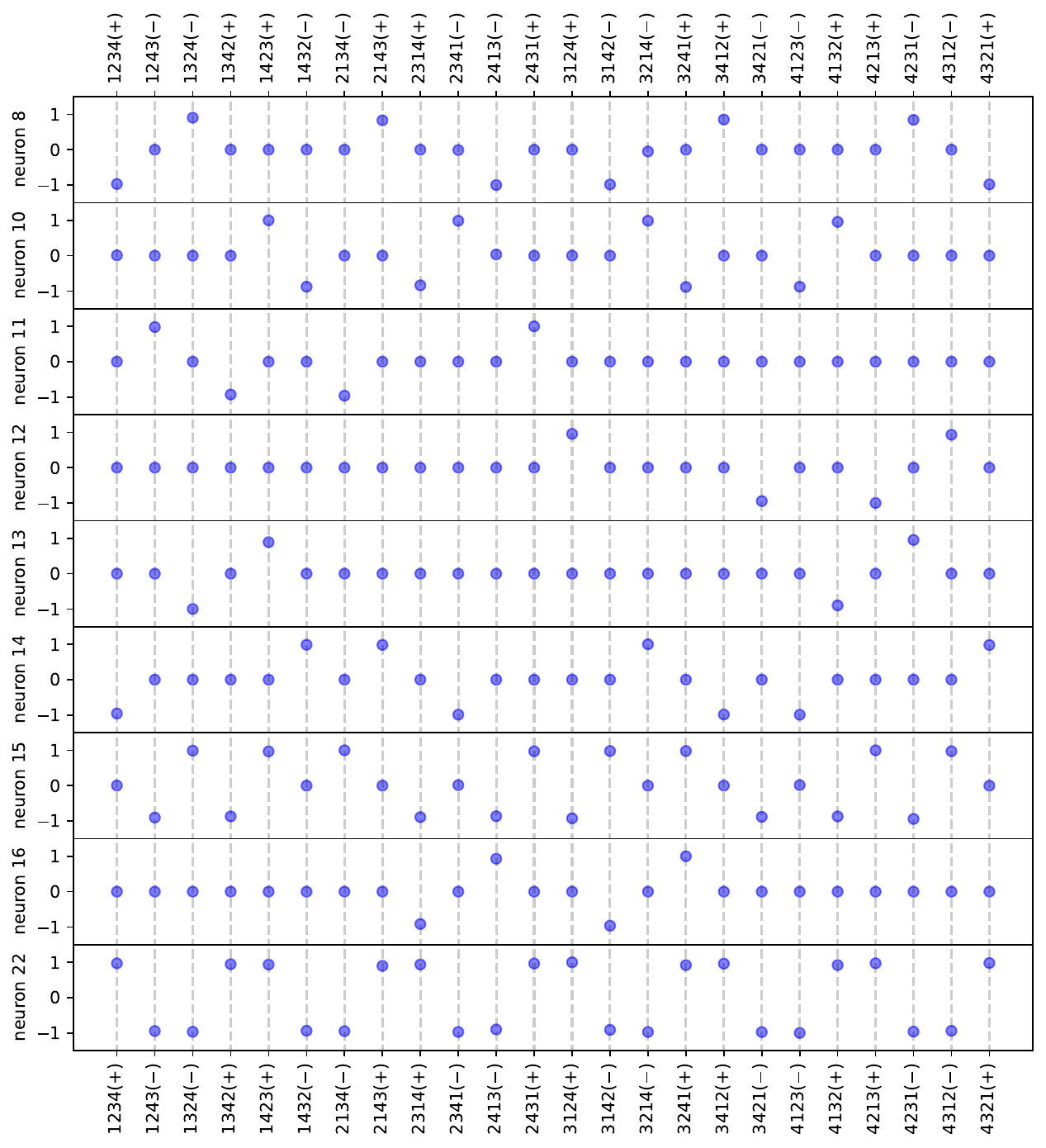}\quad
    \caption{Apply BIMT to MLP on the Permutation $S_4$ dataset. Left: the final connectivity graph, with only 9 active embedding neurons. The input layer contains embeddings of two tokens, which overlap each other but are drawn to be vertically separated. Right: the 9 active neurons correspond to group representations of $S_4$, whose values are normalized into range $[-1,1]$. In particular, neuron 22 is the sign neuron (1/-1 for even/odd permutations).}
    \label{fig:permS4}
\end{figure}

{\bf Permutation group} The task is to predict $c$ from $(a,b)$, where $a,b,c$ are elements in the 24 element group (the permutation of 4 objects) $S_4$ and $ab=c$. Our training is the same as for modular addition. Figure~\ref{fig:permS4} shows that after training with BIMT, the network is quite modular. Notice that there are only 9 active components in the embedding layer, exactly agreeing with the representation theory argument of~\cite{chughtai2023toy} ($S_4$ has a $3\times 3$ matrix representation).  In Figure~\ref{fig:permS4} (right) we show how each embedding neuron is activated by each group element, revealing that BIMT has discovered crucial group-theoretical structure! Note that we have normalized these embeddings when plotting: denote the value of the $i$-th neuron and the $j$-th token is $e_{ij}$. The normalized embedding is defined as $\tilde{e}_{ij}=e_{ij}/\left({\rm max}_j|e_{ij}|\right)$. In particular, neuron 22 is the sign neuron (1/-1 for even/odd permutations), and other active neurons correspond to subgroups or cosets (more analysis in Appendix~\ref{app:perm}).

\subsection{Extension to transformers: In context linear regression}\label{sec:exp_icl}

So far, we have demonstrated the effectiveness of BIMT for MLPs. We can generalize BIMT to transformers~\citep{vaswani2017attention}: we simply apply BIMT to linear layers in transformers (see details in Appendix \ref{app:icl}). Following the setup of~\cite{akyurek2022learning}, we now study in-context linear regression. Linear regression aims to predict $y$ from $\mat{x}\in\mathbb{R}^d$ assuming to know training data $(\mat{x}_i,\mat{y}_{i})\ (i=1,\cdots,n)$ where $\mat{y}_i=\mat{w}\cdot\mat{x}_i$. In-context linear regression aims to predict $y$ from the sequence $(\mat{x}_1,y_1,\cdots,\mat{x}_n,y_n,\mat{x})$, which is called in-context learning because the unknown weight vector $\mat{w}$ needs to be learned in context, i.e., when the transformer runs in test time rather than when it is trained. To make things maximally simple, we choose $d=1$ (the weight vector degrades to a scalar) and $n=1$.

\begin{figure}
    \centering
    \includegraphics[width=0.8\linewidth]{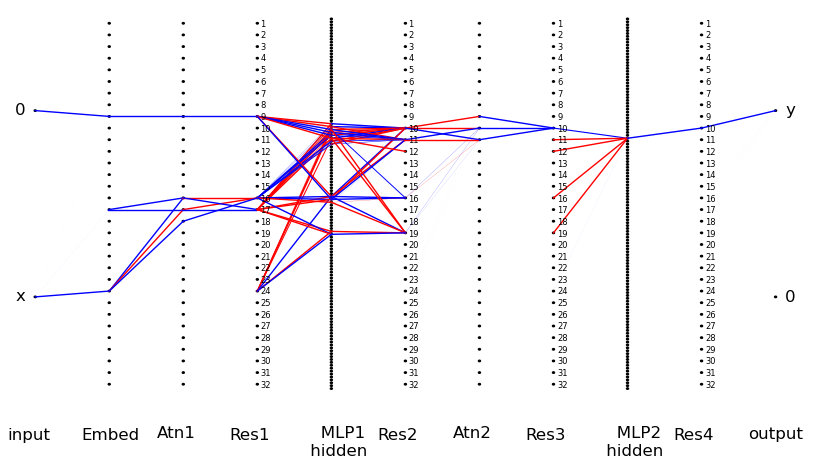}
    \includegraphics[width=0.19\linewidth]{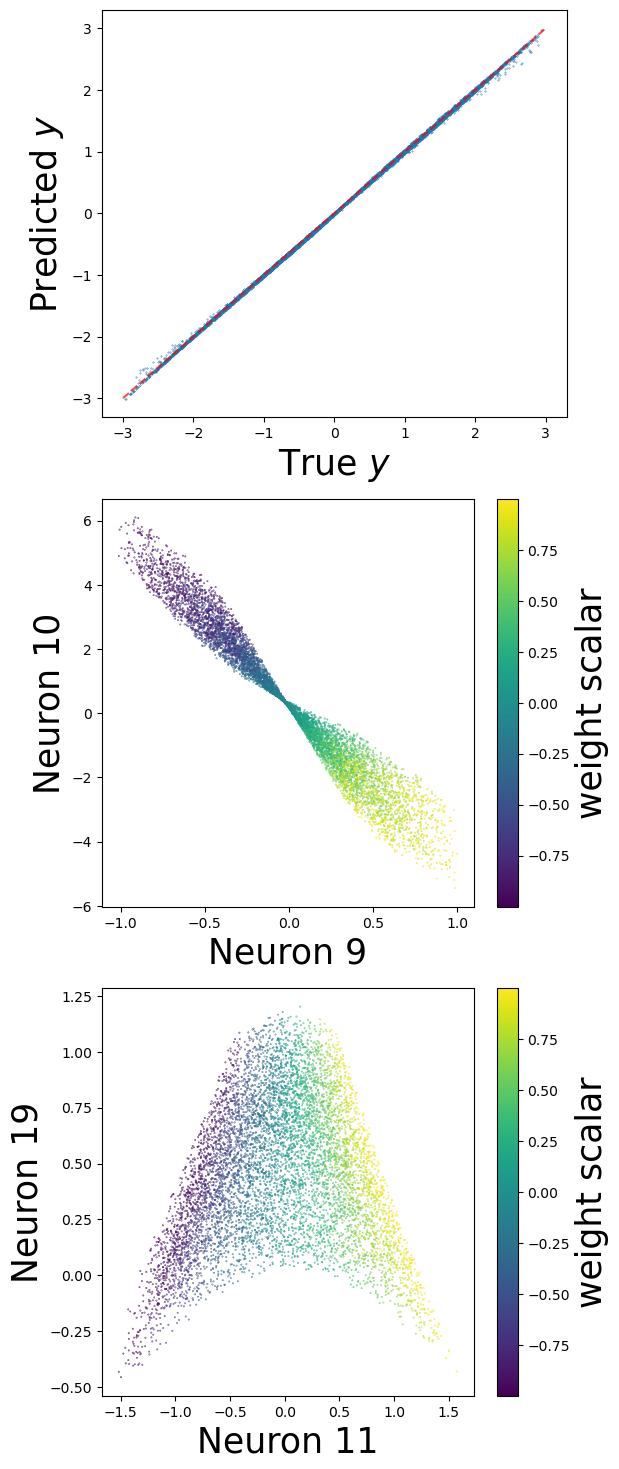}
    \caption{Application of BIMT to transformers when in-context learning linear regression. Left: the connectivity graph of the transformer after training. Only the last block is shown, which takes in $[0,x]$ to predict $[y,0]$. Right top: predicted vs true $y$. Right middle and bottom: neurons in the Res2 layer contain the information about the weight scalar, encoded non-linearly.}
    \label{fig:incontext}
\end{figure}

The architecture is displayed in Figure~\ref{fig:incontext}, where for clarity we only show the last block, ignoring its attention dependence on previous blocks. The embedding size is 32, the number of transformer layers is 2 (each layer containing an attention layer and an MLP), and the number of heads is 1. we draw $w\in U[1,3]$~\footnote{Instead we can investigate $w\sim U[-1,1]$, which has a singularity issue (please see Appendix~\ref{app:icl} for details).} and $x\in U[-1,1]$ to create datasets. With MSE loss we train with the Adam optimizer (lr: 1e-3) for $4\times 10^4$ steps ($\lambda=0.001, 0.01, 0.1, 0.3$ each for $10^4$ steps). We choose $A=2$, $y_*=0.5$, $k=30$ and $S=200$.

It is showed in ~\cite{akyurek2022learning} that $w$ is linearly encoded in neural network latent representations,  but it is not easy to track where this information is located. From Figure~\ref{fig:incontext} left, it is immediately clear which neurons are useful (active). In Figure~\ref{fig:incontext} right top, we show that the prediction is quite good even though the network has become extremely sparse. We examine active neurons in the Res2 layer, finding that several neurons are correlated with the weight scalar, although no one alone can determine the weight scalar perfectly. In Figure~\ref{fig:incontext} right middle and bottom, we show that pairs of neurons (8 and 9, 11 and 19) implicitly encode information about the weight scalar in nonlinear ways.

\subsection{Extension to tensor data: image classification}\label{sec:exp_mnist}

\begin{figure}
    \centering
    \includegraphics[width=1\linewidth]{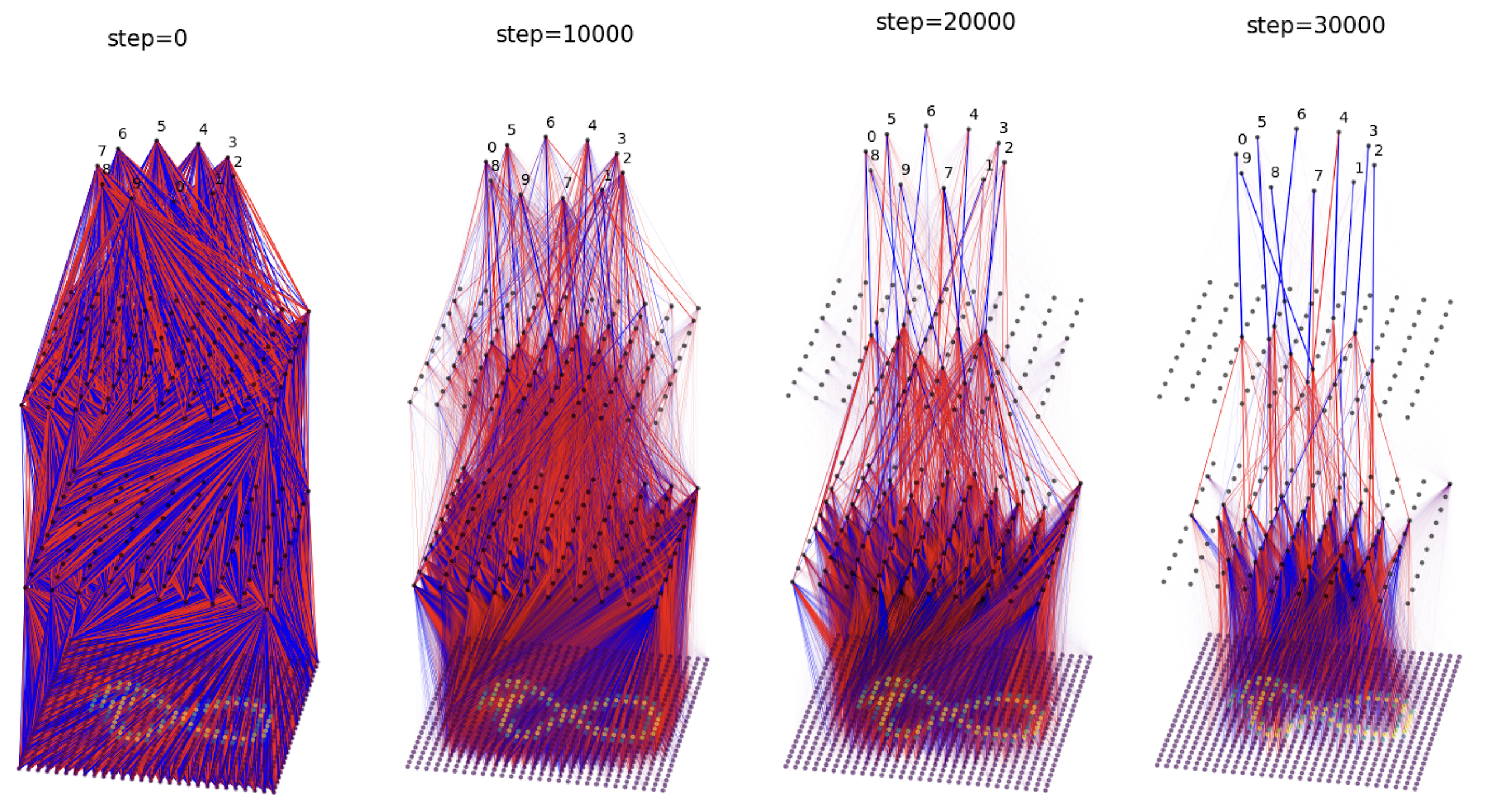}
    \caption{Application of BIMT to 3D MLP on MNIST. From left to right: connectivity graph evolution.}
    \label{fig:mnist_graph}
\end{figure}

So far, we have always embedded neural networks into 2D Euclidean space, but BIMT can be used in any geometric space. We now consider a minimal extension: embedding neural networks into a 3D Euclidean space. For 2D image data, to maintain their local structure, it is better to leave them as 2D rather than flatten them to 1D. As a result, an MLP for 2D image data should be embedded in 3D, as shown in Figure~\ref{fig:mnist_graph}. The only modification for BIMT is that when computing distances, we use 3D rather than 2D vector norms. 

We train with MSE loss and use the Adam optimizer (lr=1e-3) for $4\times 10^4$ steps ($\lambda=0.001, 0.01, 0.1, 0.3$ each for $10^4$ steps). We choose $A=2$, $y_*=0.5$, $k=30$ and $S=200$. We disable swaps of input pixels. We show the evolution of the network in Figure~\ref{fig:mnist_graph}. Starting from a dense network, the network becomes more modular and sparser over time. Notably, the receptive field shrinks for the input layer, since BIMT learns to prune away peripheral pixels which always equal zero. Another interesting observation is that most of the weights in the middle layer are negative (colored red), while most of the weights in the last layer are positive (colored blue). This suggests that the middle layer is not adopting the strategy of pattern matching, but \textit{pattern mismatching}. Pattern matching/mismatching means: if an image has/does not have these patterns, it is more likely to be an 8, say. We visualize features in Appendix \ref{app:mnist}, where we also include the results for MLPs with different depths. Moreover in the output layer, class 1 and 7 are automatically swapped to become neighbors, probably due to their similarity. In future works we would like to compare our method with convolutional neural networks (CNN). It might be best to combine CNN with BIMT, since CNN guarantees the locality of inputs, while BIMT encourages locality of model internals. 

\section{Related Work}\label{sec:related_works}

{\bf Mechanistic Interpretability} (MI) is an emerging field that aims to mechanically understand how neural networks work. Various modules/circuits are identified from neural networks via reverse engineering, including image 
circuits~\citep{olah2020zoom}, induction heads~\citep{olsson2022context}, computational quanta~\citep{michaud2023quantization}, transformer circuits~\citep{elhage2021mathematical}, factual associations~\citep{meng2022locating} and heads in the wild~\citep{wang2023interpretability}, although superposition~\citep{elhage2022superposition} makes interpretability more complicated. A generalization puzzle called grokking~\citep{power2022grokking} has also been understood by reverse engineering neural networks~\citep{nanda2023progress, chughtai2023toy, liu2023omnigrok,liu2022towards}.

{\bf Modularity} in neural networks can help generalization in transfer learning~\citep{pfeiffer2023modular}, as well as enhance interpretability~\citep{olah2020zoom}. Non-modular neural networks trained in standard ways are shown to present some yet imperfect extent of  modularity~\citep{filan2021clusterability, hod2021detecting,csordas2021are}. Modular networks explicitly use trainable modules in constructing neural networks~\citep{kirsch2018modular,azam2000biologically}, but this inductive bias may require prior knowledge about the tasks. The multi-head attention layer in transformers lies in the category of explicitly introducing modularity. By contrast, this work does not explicitly introduce modules, but rather lets modules emerge from non-modular networks with the help of BIMT.

{\bf Pruning} can lead to sparse and efficient neural networks~\citep{han2015learning, anwar2017structured, blalock2020state, frankle2018lottery}, usually achieved by $L_1$ or $L_2$ regularization and thresholding small weights to zero. BIMT borrows the $L_1$ regularization technique for sparsity, but improves modularity by making the $L_1$ regularization distance-dependent.

{\bf Analogy between neuroscience and neural networks} has existed for long in the literature~\citep{richards2019deep, hassabis2017neuroscience}. Although biological and artificial neural networks may not have the same low-level learning mechanisms~\citep{lillicrap2020backpropagation}, we can still borrow high-level ideas and concepts from neuroscience to design more interpretable artificial neural networks, which is the goal of this work. The minimal connection cost idea has been explored in~\cite{clune2013evolutionary, mengistu2016evolutionary,huizinga2014evolving,ellefsen2015neural}, where an evolutionary algorithm is applied to evolve tiny networks. By contrast, our method is more aligned with modern machine learning, i.e., gradient-based optimization and broader applications.

\section{Conclusions and Discussion}\label{sec:conclusions}

We have proposed brain-Inspired modular training (BIMT), which explicitly encourages neural networks to be modular and sparse. BIMT is a principled idea that could generalize to many types of data and network architectures. Tested on several relatively small-scale tasks, we show its ability to give interpretable insights for these problems. In future studies, we would like to see if this training strategy remains valid for larger-scale tasks, e.g., large language models (LLM). In particular, can we fine tune LLMs with BIMT to make them more interpretable? Moreover, BIMT achieves interpretability at the price of slight performance degradation. We would like to improve BIMT such that interpretability and performance are achieved at the same time.

{\bf Broader Impacts} We believe that building interpretable neural networks will make AI more controllable, more reliable and safer. However, like other AI interpretability research, the controllability brought by interpretability should be regulated, making sure the technology is not misused.

{\bf Limitations} This work deals with small-scale toy problems, where neural networks can be easily visualized. It is still unclear whether this method remains effective for larger-scale problems.


\bibliography{ref.bib}
\bibliographystyle{plainnat}

\clearpage

\begin{center}
    {\LARGE\bf  Supplementary material}
\end{center}

\begin{appendix}

\section{Pruning}\label{app:pruning}

\begin{figure}[htbp]
    \centering
    \includegraphics[width=0.5\linewidth]{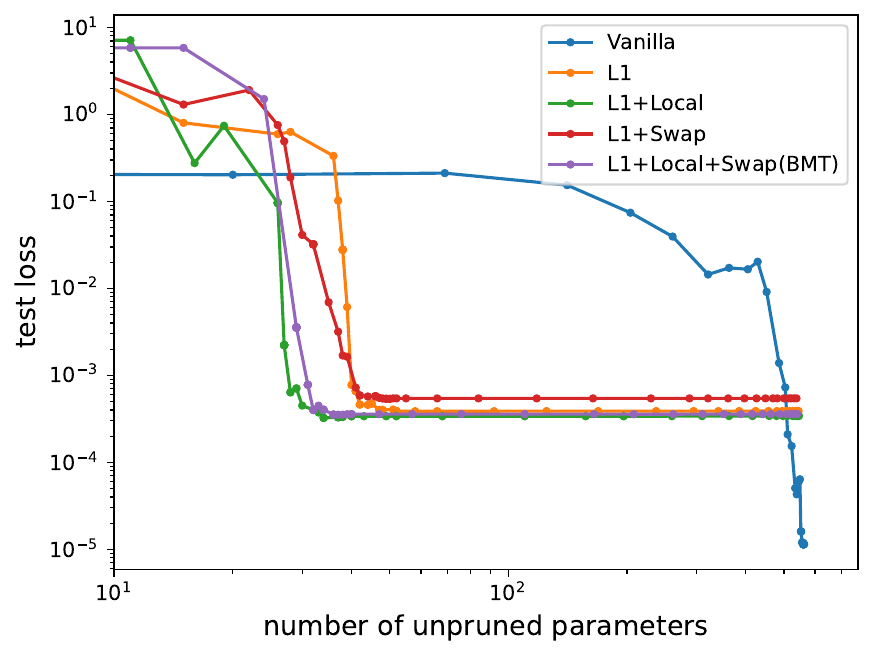}
    \caption{In a toy regression problem (see Figure~\ref{fig:sf_method}), BIMT and L1 + Local achieve the best pruning results, better than L1 regularization alone (a standard training method for pruning).}
    \label{fig:sf_ablation_prune}
\end{figure}

Although the original goal of BIMT is to make neural networks modular and interpretable, the fact that it can make networks sparse is also useful for pruning. Here we show the benefits of BIMT in terms of pruning on a toy example (in Figure~\ref{fig:sf_method}). The task is to fit $(y_1,y_2)=(x_1x_4+x_2x_3, x_1x_4-x_2x_3)$ from $(x_1,x_2,x_3,x_4)$ with a two-hidden-layer MLP. As in Figure~\ref{fig:sf_method}, we test five training methods: vanilla, L1, L1 + Local, L1 + Swap and BIMT (L1 + Local + Swap). For each trained network, we sort their parameters (including weights and biases) from small to large (in magnitudes), defining a threshold below which parameters are set to zero. Given a threshold, we can compute the number of unpruned parameters $N_u$, as well as test loss $\ell_{\rm test}$. By sweeping the threshold, we obtain a tradeoff frontier, as shown in Figure~\ref{fig:sf_ablation_prune}. Note that in this plot, the lower left the curve goes, the better the pruning. So BIMT and L1 + Local achieve the best pruning results, better than L1 (which is standard in pruning). We leave the full investigation of BIMT as a pruning method in future works.

\section{Symbolic formulas}\label{app:sf}

\subsection{How good are the predictions?}
Since the connectivity graphs in Figure~\ref{fig:sf_various} are extremely sparse, one may suspect that these sparse networks severely underfit. We show that this is not the case, since the (test) losses are quite low, and sample-wise prediction errors are quite small, as shown in Figure~\ref{fig:sf_accurate}. The explanation is that SiLU activations (and other similar activations) are surprisingly effective. In fact, in the following we reverse engineer how these symbolic functions can be approximated with very few parameters.

\begin{figure}
    \centering
    \includegraphics[width=0.3\linewidth]{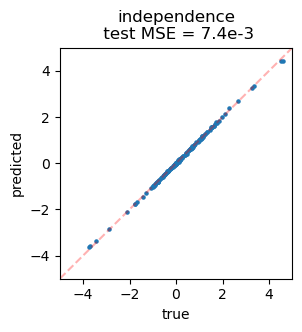}
    \includegraphics[width=0.32\linewidth]{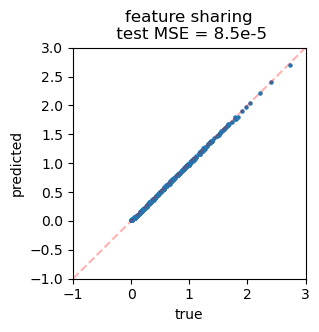}
    \includegraphics[width=0.32\linewidth]{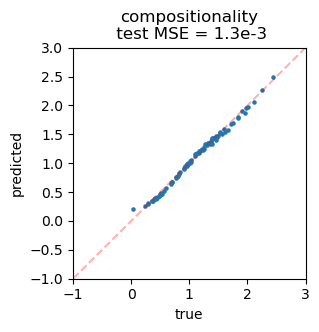}
    \label{fig:sf_accurate}
\caption{Although the networks are extremely sparse in Figure~\ref{fig:sf_various}, their predictions are quite good.}
    \label{fig:sf_accurate}
\end{figure}

\subsection{Reverse engineering how the formulas are implemented}

Once we obtain the sparse connectivity graphs in Figure~\ref{fig:sf_various}, we can easily reverse engineer how neural networks approximate these analytical functions with linear operations and SiLU activation functions $\sigma(x)=x/(1+e^{-x})$. If not stated otherwise, the approximations below hold for $x\in[-1,1]$.

{\bf (a) Independence}
\begin{equation}
\begin{aligned}
    x^2&\approx -1.33x + 1.84\sigma(1.53x), \\
    {\rm sin}(x) &\approx  -2.27x + 1.72\sigma\left[-0.91\sigma(-3.24x+1.54)+2.63\right] - 2.10, \\
    x^3 &\approx 2.30\sigma\left[3.34\sigma(0.90x-0.51)-0.46\right]-2.27\sigma\left[3.00\sigma(-0.87x-0.19)-1.07\right].
\end{aligned}
\end{equation}

{\bf (b) feature sharing}
\begin{equation}
    x^2 \approx 0.35\sigma\left[1.41\sigma(2.64x)+1.99\sigma(-1.80x+0.05)\right].
\end{equation}

{\bf (c) compositionality}
\begin{equation}
    1.60\sqrt{x-1.24}\approx 0.80\sigma(1.04x) - 1.18\sigma\left(-2.26x+2.44\right) - 0.18,\quad x\in[1.24,3.66].
\end{equation}

\subsection{Intermediate quantities}
\begin{figure}[htbp]
    \centering
    \includegraphics[width=0.5\linewidth]{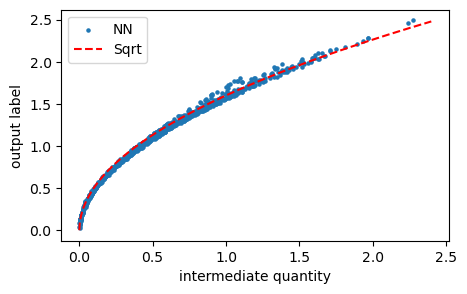}
    \caption{Verifying the existence of an intermediate quantity.}
    \label{fig:intermediate_compare}
\end{figure}

In the compositionality example, $y=\sqrt{(x_1-x_2)^2+(x_3-x_4)^2}\equiv \sqrt{I}$, we argue that there is an intermediate quantity $I$ contained in the network. Figure~\ref{fig:intermediate_compare} shows this to be neuron 11 in A3. The relation between this neuron and the NN output is seen to accurately approximate the square root function.

\section{Two moon classification}\label{app:two_moon}

\begin{figure}[htbp]
    \centering
    \includegraphics[width=1\linewidth]{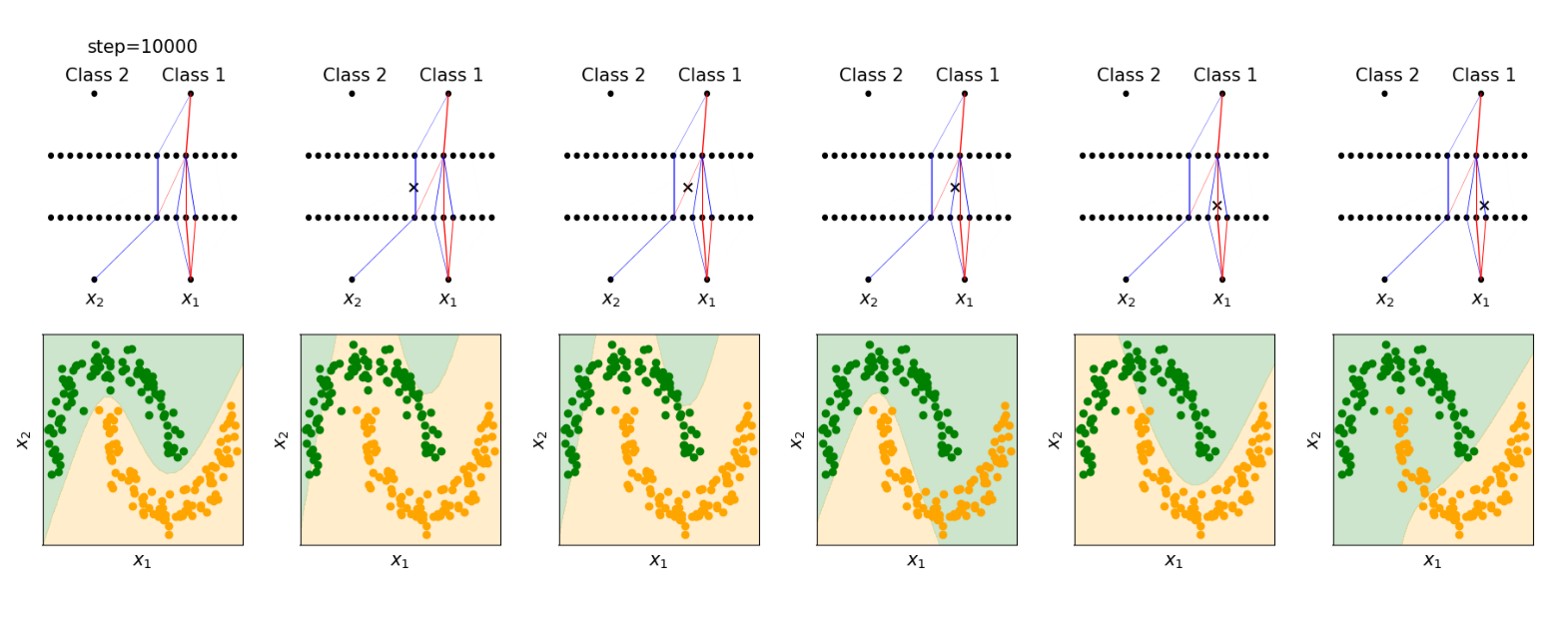}
    \caption{For the two moon dataset, we interpret what each weight is doing by setting it to zero (marked with a cross in the top panel) and visualizing the resulting decision boundary (bottom).}
    \label{fig:two_moon_intervene}
\end{figure}

We showed in Section \ref{sec:exp_two_moon} that a very sparse network is able to classify the two moon datasets. Since the active weights are so few, we are able to interpret each of them by removing it (setting the weight value to be 0). We show how the decision boundaries change under removing one of the weight (marked as a cross) in Figure~\ref{fig:two_moon_intervene}. It is clear that every weight is necessary for prediction, since removing any of them can lead to false classifications. We can also write down the symbolic formula for the network ($\sigma(x)=x/(1+e^{-x})$)
\begin{equation}
    \begin{aligned}
        p({\rm green}|x_1,x_2)  = & \frac{{\rm exp}(s(x_1,x_2))}{1+{\rm exp}(s(x_1,x_2))} \\
        s(x_1,x_2) = & 5.16\sigma(1.44x_2+1.43) - 6.36\sigma(-0.86\sigma(1.44x_2+1.43)+1.72\sigma(1.34x_1)
        \\ 
        &-2.47\sigma(-3.29x_1-0.17)+1.99\sigma(2.32x_1-2.07)).
    \end{aligned}
\end{equation}

\section{Algorithmic datasets}\label{app:modadd_sensitivity}

\begin{figure}[htbp]
    \centering
    \includegraphics[width=1\linewidth]{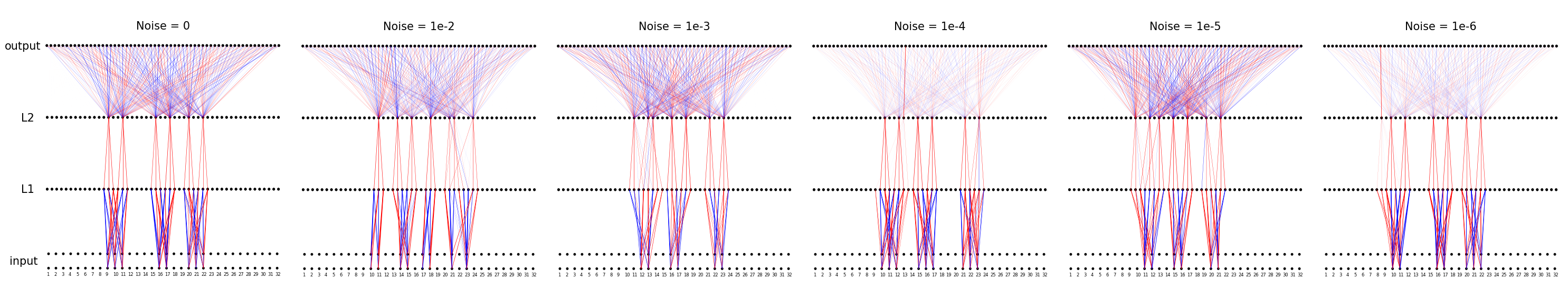}
    \includegraphics[width=1\linewidth]{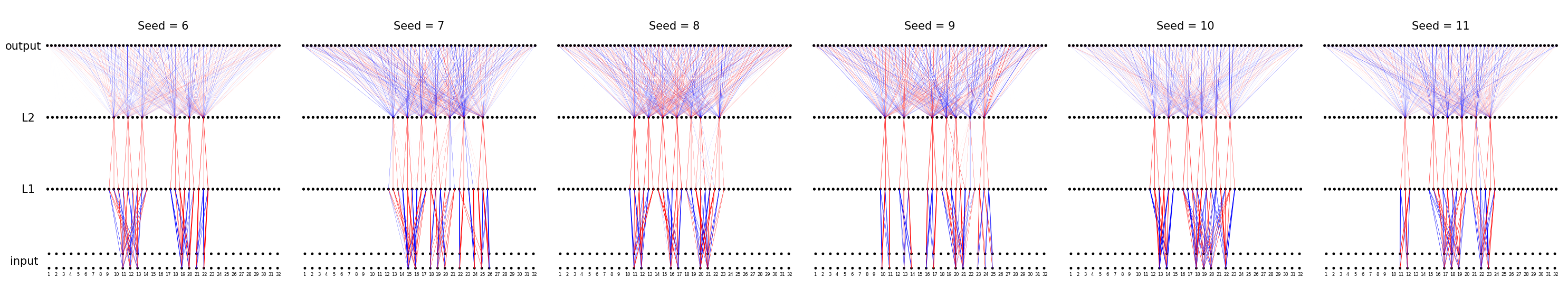}
    
    \caption{For the modular addition task, we test what happens when we add small perturbations to the model (top), and when we initialize the the parameters using different random seeds (bottom).}
    
    \label{fig:noise_seeds}
\end{figure}

{\bf Sensitivity (add noise)}
In the modular addition task in Section \ref{fig:modadd}, we investigate the sensitivity of the module structures to small perturbations during their initializations. To do this, we first initialize a model's parameters using a fixed random seed, and then add zero-mean Gaussian noise with varying standard deviations $\sigma$. In Figure~\ref{fig:noise_seeds} top presented in the graphs, the "noise" values refer to the standard deviation of the Gaussian noise added to the model's parameters during initialization.

We find that small perturbations to initialization has sizeable impacts on the final model. Even the least perturbed model ($\sigma=10^{-6}$) is quite different from the base model going from layer 2 to the output, although the modules are mostly in the same positions. We conjecture that training dynamics has many branching points, where a small perturbation can lead to quite different basins. Luckily, these basins all have similar tree structures, amenable to be interpreted.

{\bf More tree graphs} We also investigate the behavior of the model with different random seeds for initialization and see a diverse pattern of module formation. Although they are different in details, there are some universal features: (1) the number of modules is odd the most times, supporting the argument of (majority) voting; (2) between layer 1 and layer 2, many copies of the same motif emerges, which connects three neurons in L1 to one neuron in L2.

\section{Reverse engineering learned $S_4$ embeddings}\label{app:perm}

\subsection{Visualizing neurons with the Cayley graphs}

\begin{figure}[htbp]
    \centering
    \includegraphics[width=1\linewidth]{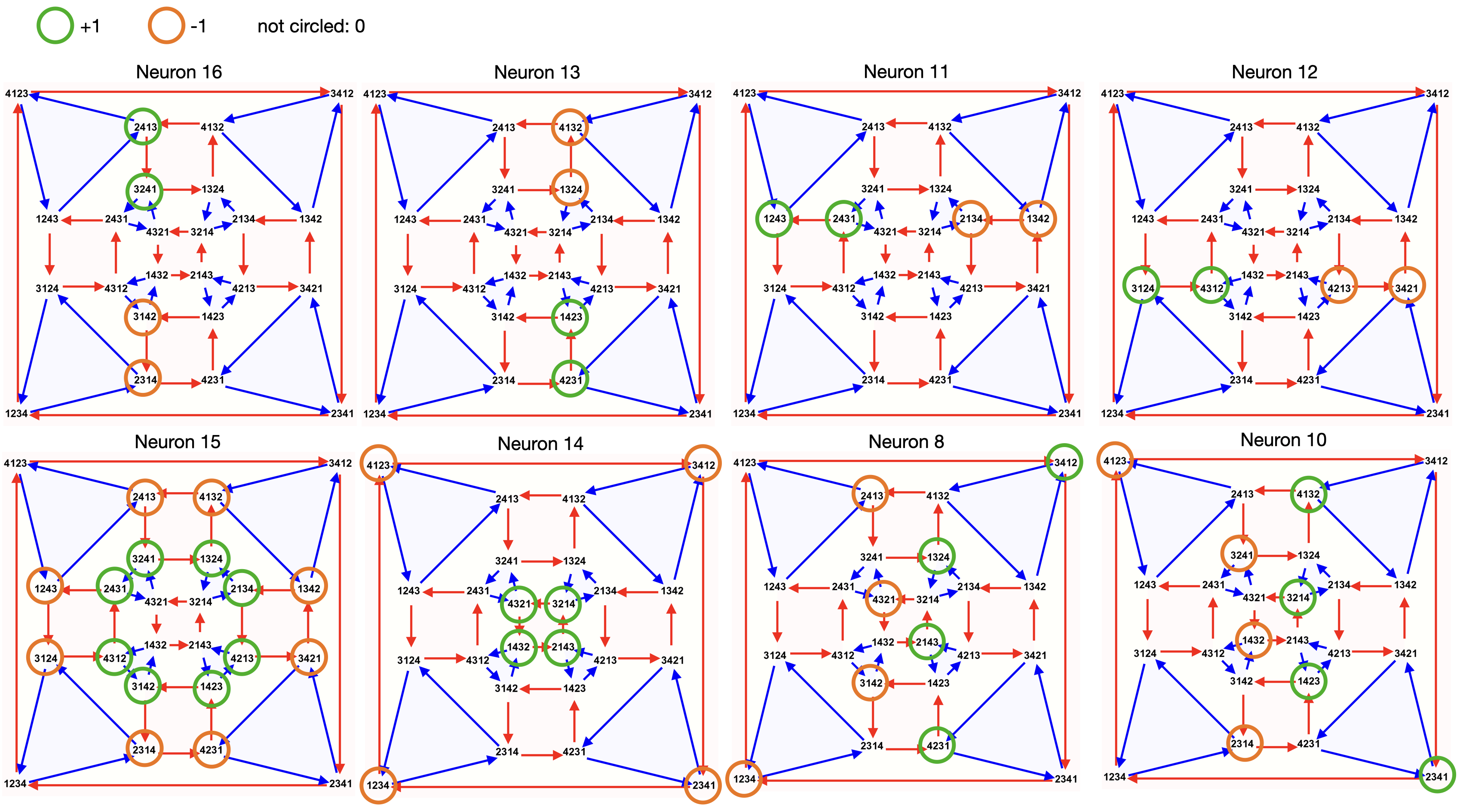}
    \caption{For active neurons in Figure~\ref{fig:permS4}, we highlight active group elements on Cayley graph (green/orange/no circle means 1/-1/0), revealing interesting structure.}
    \label{fig:S4_cayley}
\end{figure}

In Figure~\ref{fig:permS4}, we find that there are only 9 active embedding neurons. For each of them except for the sign neuron, only a subset of group elements is non-zero, and interestingly, non-zero elements are close to +1 or -1. So to visualize what each neuron is doing, we can highlight its active group elements on a Cayley graph of the permutation group $S_4$, as shown in Figure~\ref{fig:S4_cayley}, where green/orange/no circle means +1/-1/0, respectively. Red and blue arrows represent two generators 4123 and 2314. There are a few interesting observations: (1) By moving circles along blue arrows for neuron 8 gives neuron 10; (2) For neuron 14 (15), the inside square (octagon) activates to +1, while the outside square (octagon) activates to -1. Moreover, both of them are closed under red arrows. (3) For neuron 11, 12, 13, 16, they display similar structures, up to translations and rotations.

\subsection{The learned embedding is \textit{not} linear transformation of the faithful group representation}

Although a lot of interesting structure emerges from learned embeddings, we show that the learned embedding is not linear transformation of the faithful group representation; at least some extent of non-linearity is at play.

$S_4$ has a $3\times 3$ (truthful) matrix representation, corresponding to 3D rotations and reflection of the tetrahedron. We denote this representation as $\mat{E}^{\rm true}_i\in\mathbb{R}^{3\times 3}$, and we denote the learned embedding $\mat{E}_i\in\mathbb{R}^9 (i=1,\cdots, 24)$. If $\mat{E}_i^{\rm true}$ and $\mat{E}_i$ are linearly related, there exists $\mat{A}\in\mathbb{R}^{3\times 3}$ and $\mat{V}\in\mathbb{R}^{9\times 9}$ such that
\begin{equation}
    \mat{E}_i = \mat{V}{\rm vec}(\mat{A}\mat{E}_i^{\rm true}\mat{A}^{-1}),
\end{equation}
where ${\rm vec}$  flattens a matrix to a vector. We define a loss function:
\begin{equation}
    L(\mat{V},\mat{A}) = \frac{\sum_{i=1}^{24} |\mat{E}_i - \mat{V}{\rm vec}(\mat{A}\mat{E}_i^{\rm true}\mat{A}^{-1})|^2}{\sum_{i=1}^{24} |\mat{E}_i|^2}.
\end{equation}
If $L\approx 0$, this means a linear relation between $\mat{E}_i^{\rm true}$ and $\mat{E}_i$, otherwise nonlinearity is present. We optimize the above loss function with \texttt{scipy.optimize.minimize}, consistently finding the minimal value to be 0.56 (same for 100 random seeds), implying that learned embedding $\mat{E}_i$ is not a linear transformation of a truthful group representation.

Since the learned representation is quite sparse, probably no single truthful representation can reach that sparsity (defined below). We conjecture that the learned representation could be combining multiple (sparse) representations in a clever way such that the combined representation is even more sparse, and remains "faithful" to the extent that prediction accuracy is perfect. The "combination" might not be that surprising, since we saw that for modular addition (Figure~\ref{fig:modadd}), the learned embedding consists of three different faithful group representations.  

To measure sparsity of a representation, we define a representation matrix $\mat{R}$ and its normalized version $\tilde{\mat{R}}$:
\begin{equation}
    \mat{R} \equiv [\mat{E}_1,\cdots, \mat{E}_{24}]\in\mathbb{R}^{9\times 24},\quad  \tilde{\mat{R}} \equiv \frac{\mat{R}}{|{\rm vec(\mat{R})|_1}},
\end{equation}
and its entropy $S$ and effective dimension $D$ as
\begin{equation}
    S \equiv {\rm Entropy}({\rm vec}(\tilde{|\mat{R}|})), \quad D\equiv 2^S.
\end{equation}
For faithful representations corresponding to tetrahedra
\begin{equation}
\begin{aligned}
    &A: (1,0,0), (-1,1,0), (0,-1,1), (0,0,-1) \\
    &B: (-1,0,0), (0,-1,0), (0,0,-1), (1,1,1) \\
    &C: (1,-\frac{1}{\sqrt{3}},-\frac{1}{\sqrt{6}}), (-1,-\frac{1}{\sqrt{3}},-\frac{1}{\sqrt{6}}),
    (0,\frac{2}{\sqrt{3}},-\frac{1}{\sqrt{6}}),
    (0,0,\frac{\sqrt{6}}{2}), 
\end{aligned}
\end{equation}
their effective dimensions are $D\approx 120, 108, 153$, respectively, while the learned representation has $D\approx 80$, which is noticeably smaller.

\section{In-context learning}\label{app:icl}
In this section, We show how to modify BIMT (presented in Section \ref{sec:method} for MLPs) to use with transformers.

\subsection{Applying BIMT to transformers} 

In Section~\ref{sec:method}, we discussed how to do BIMT with fully-connected neural networks; generalization to transformers is also possible: we simply apply BIMT to "linear layers", which include not only linear layers in MLPs, but also (key, query, value) matrices, embed/unembed layers, as well as projection layers in attention blocks. In summary, we count any matrix as a "linear layer" if the matrix belongs to model parameters and does matrix-vector multiplications.


{\bf Attention layers} can be seen as a special type of linear layers, involving $[\mat{W_Q}, \mat{W_K}, \mat{W_V}]$ as the weight matrix. We leave softmax and dot product of keys and queries unchanged, since they involve no trainable parameters. The way to calculate regularizations is the same as MLPs. However, care needs to be taken when swapping neurons. We want to swap neurons (with their corresponding weights and biases) such that the whole network remains unchanged as a function. For MLPs, we can therefore swap any two neurons in the same layer (and their corresponding weights and biases). However for transformers, since each head operates independently, only neurons in the same head can be swapped. In addition, two heads in the same attention layer can be swapped. In summary, swapping choices are more restricted for attention layers.

{\bf Residual connections} For MLPs, swapping can be implemented independently for each layer. However, the residual connections couple all the layers on the residual stream. This means that all layers on the residual stream share the same permutations/swapping. 

{\bf LayerNorm} normalizes features, which contradicts the goal of sparsity. Currently we simply remove LayerNorm layers, which works fine for our two-layer transformers. In the future, we would like to explore principled ways to handle LayerNorm in the framework of BIMT.

\subsection{A singularity problem}

In Section \ref{sec:exp_icl}, we trained a transformer with BIMT for an in-context learning linear regression problem. Out setup simply has $d=1$ and $n=1$, which means that given $(x_1,y_1,x)$ and knowing $y_1=wx_1$, the network aims to predict $y=wx$ based on $x_1,x,y_1$. The ground truth formula is $y = \frac{y_1}{x_1}x$, which is singular at $x_1=0$. In Section \ref{sec:exp_icl}, we explicitly constrain $x_1$ to be positive and bound it away from zero, to avoid the possible singularity. In this section, we investigate the effect of the singularity. The setup is exactly the same as in Section \ref{sec:exp_icl}, with the only difference that $x_1, x$ are now drawn from $U[-1,1]$ instead of $U[1,3]$, where $U[a,b]$ stands for a uniform distribution on $[a,b]$. 

\begin{figure}
    \centering
    \includegraphics[width=0.8\linewidth]{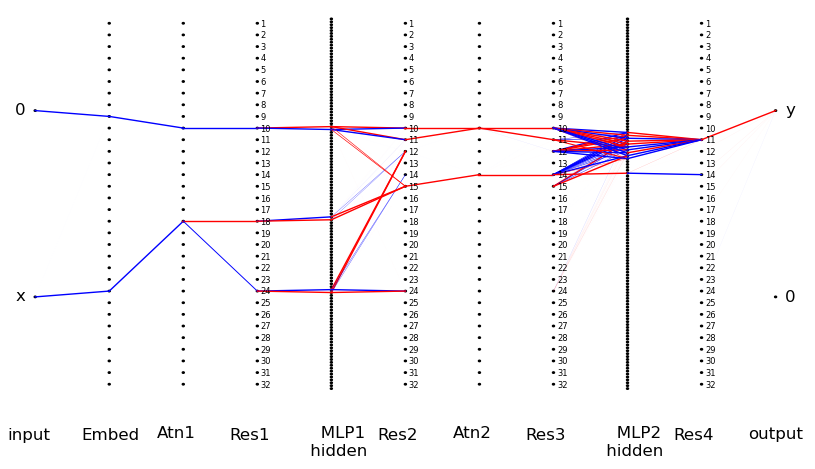}
    \includegraphics[width=0.19\linewidth]{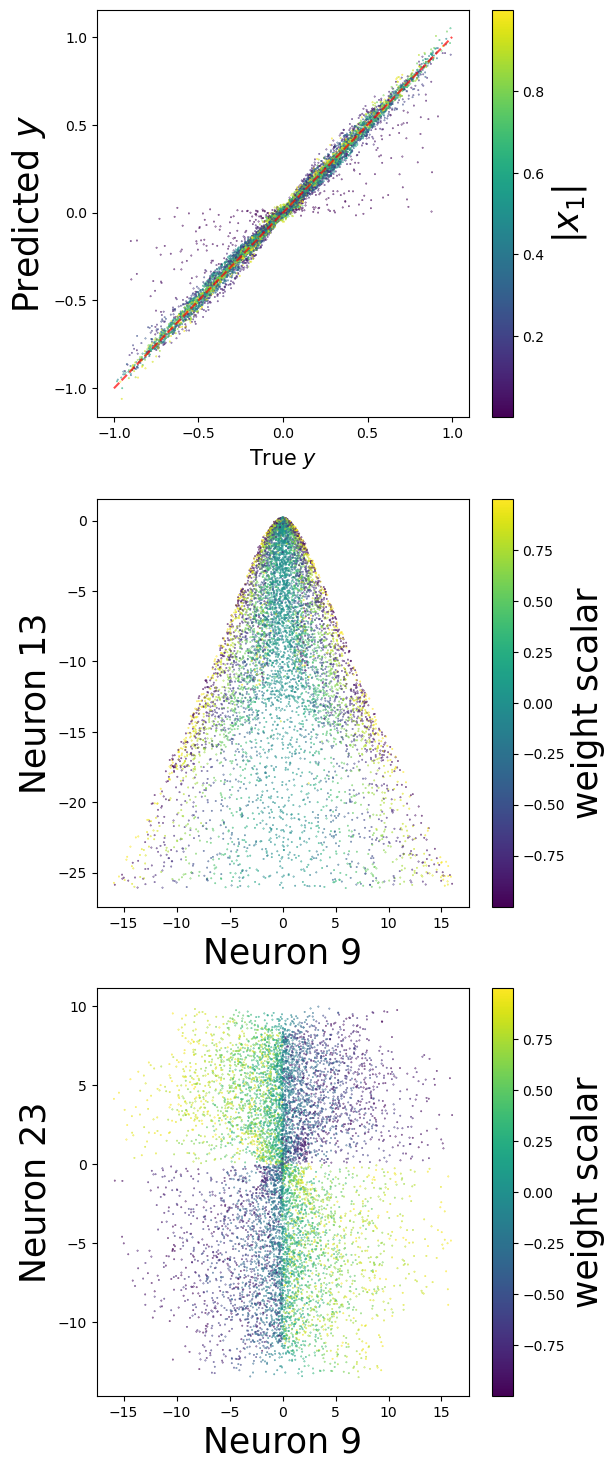}
    \caption{Apply BIMT to transformers on in-context learning linear regression. The setup is almost the same as in Figure~\ref{fig:incontext}, except that here data present some singularities.}
    \label{fig:incontext_singular}
\end{figure}

After training with BIMT, the transformer is shown in Figure~\ref{fig:incontext_singular}. The right top shows the predicted $y$ versus the true $y$, where the prediction is good for large $|x_1|$ and bad for small $|x_1|$, which is an indication of the singularity point. Moreover, similar to in Section \ref{sec:exp_icl}, we look for neurons that potentially encode the information of the weight scalar in the Res2 layer. We find that neurons 9, 13, 23 are correlated with the weight scalar, although none of them can predict the weight scalar single-handedly. In Figure~\ref{fig:incontext_singular} right bottom, the 2D plane, spanned by neuron 9 and neuron 23, is split into four regions, with very abrupt changes on the boundaries, which are also evidence for the singularity.

\section{MNIST}\label{app:mnist}

{\bf Applying BIMT to tensor data} For simplicity, we usually embed neural networks in 2D Euclidean space, but it can be any geometric space. For image data, for example, to maintain locality of input images, it is more reasonable to embedd neural networks into 3D Euclidean space (2 along image axes, 1 along depth). Now neurons in the same layer are arranged in a 2D grid instead of a 1D grid. This only affects distances between neurons, with everything else unchanged. In fact, to change MLP embedding from 2D to 3D, the only thing we need to change is to redefine the coordinates of neurons. Similarly, networks can be embedded in higher-dimensional Euclidean space or even Riemannian manifolds, by properly redefining coordinates and computing distances based on the manifold metric.

\begin{figure}
    \centering
    \includegraphics[width=1\linewidth]{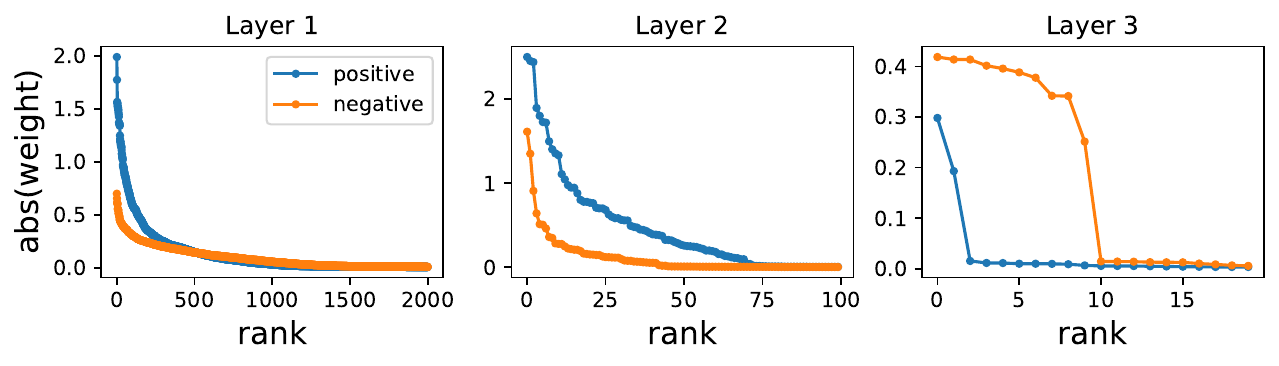}
    \caption{The magnitudes of positive and negative weights in MLP layers after training. Positive weights dominate in Layer 2, while negative weights dominate in Layer 3.}
    \label{fig:mnist_weights}
\end{figure}

{\bf Positive vs negative weights} It was observed in Figure~\ref{fig:mnist_graph} that at the end of training, most weights in layer 3 (the last layer before outputs) are positive (blue), while most weights in layer 2 are negative (red). To verify that this is not just a visual artifact, we plot the rank distribution of positive and negative weights in Figure~\ref{fig:mnist_weights}. In Layer 1, there are more positive weights large in magnitude, while negative weights seem to have a heavier tail. In Layer 2 and Layer 3, there are clearly more positive and negative weights, respectively. We are still not sure why such symmetry breaking happens, because at initializations, the number of positive and negative weights are roughly balanced. In Section~\ref{sec:exp_mnist}, we called this phenomenon "pattern mismatching". It would be interesting to investigate if pattern mismatching is prevalent in neural networks, or is specific to some combinations of specific architectures, datasets and/or training techniques.

\begin{figure}
    \centering
    \includegraphics[width=0.7\linewidth]{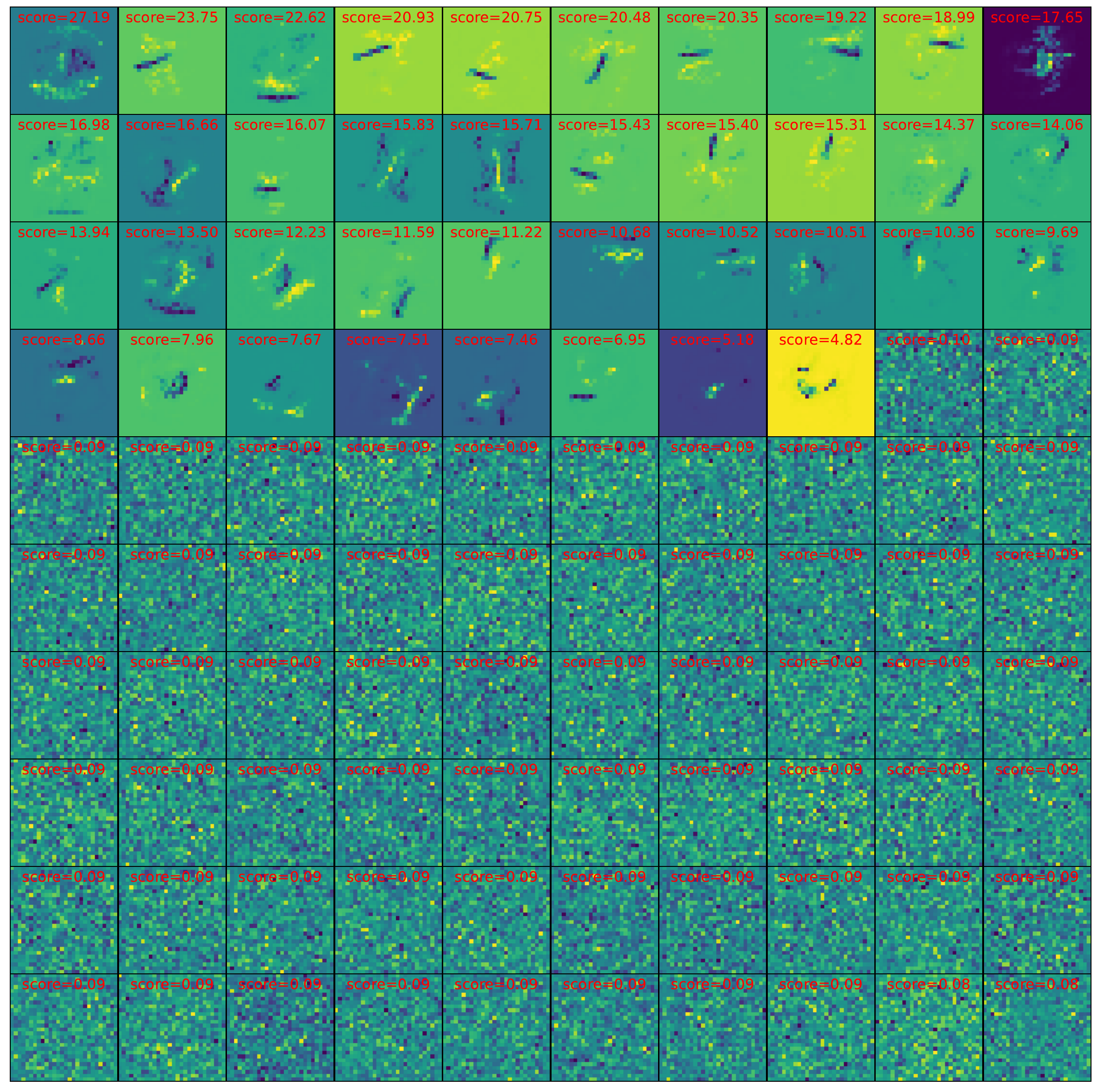}
    \caption{Visualizing MNIST features (Layer 1 of Figure~\ref{fig:mnist_graph}).}
    \label{fig:mnist_feature}
\end{figure}

{\bf Learned features} To understand what the neural network has learned, we visualize the features (weight matrices) in Layer 1. For each feature, we compute its score as sum of absolute weights. We rank features from high to low scores, finding there are 38 features with large scores, as shown in Figure~\ref{fig:mnist_feature}.The features look like intermediate to high level feature maps of convolutional filters in trained convolutional neural networks, since they are more than just edge detectors (low-level convolutional filters), containing some extent of global correlations.

\begin{figure}
    \centering
    \includegraphics[width=1\linewidth]{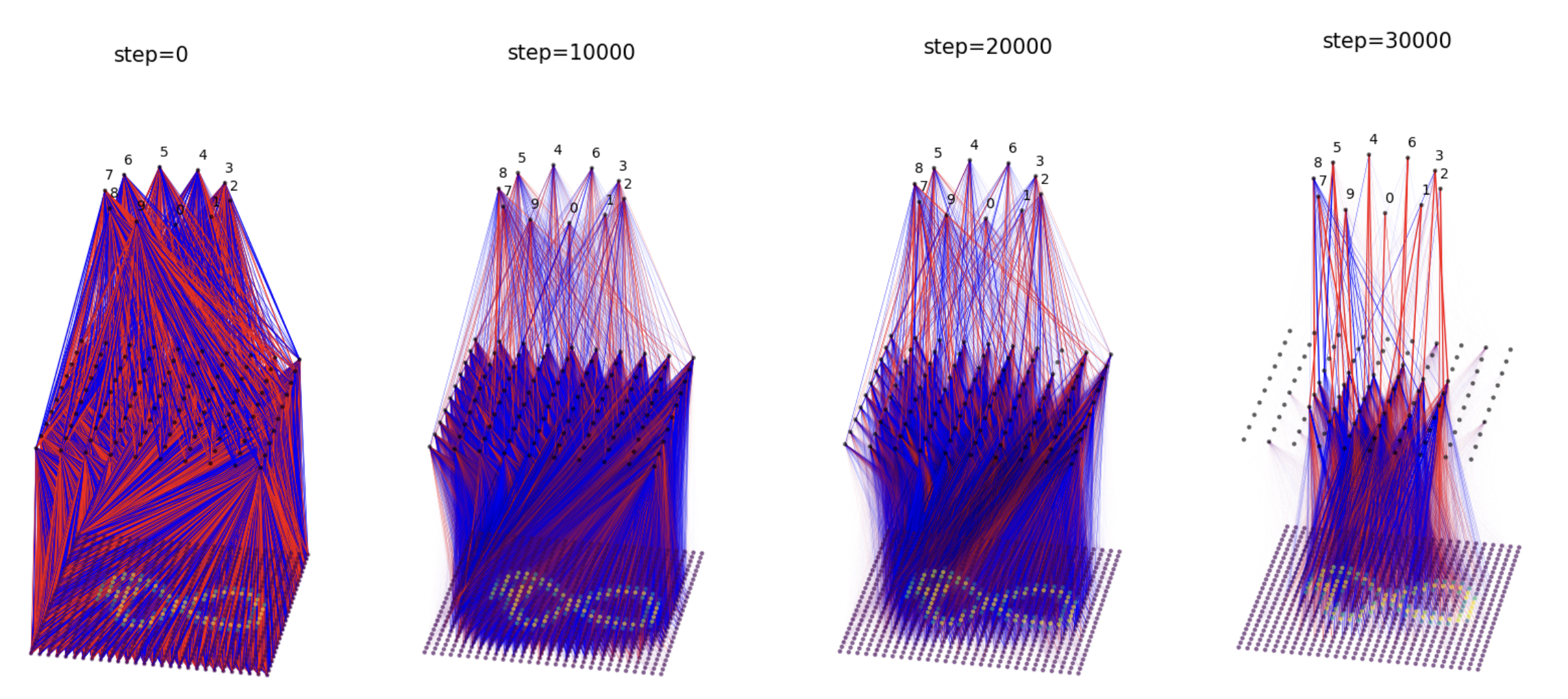}
    \caption{Evolution of a 2 Layer MLP trained with BIMT.}
    \label{fig:mnist_2L}
\end{figure}

\begin{figure}
    \centering
    \includegraphics[width=1\linewidth]{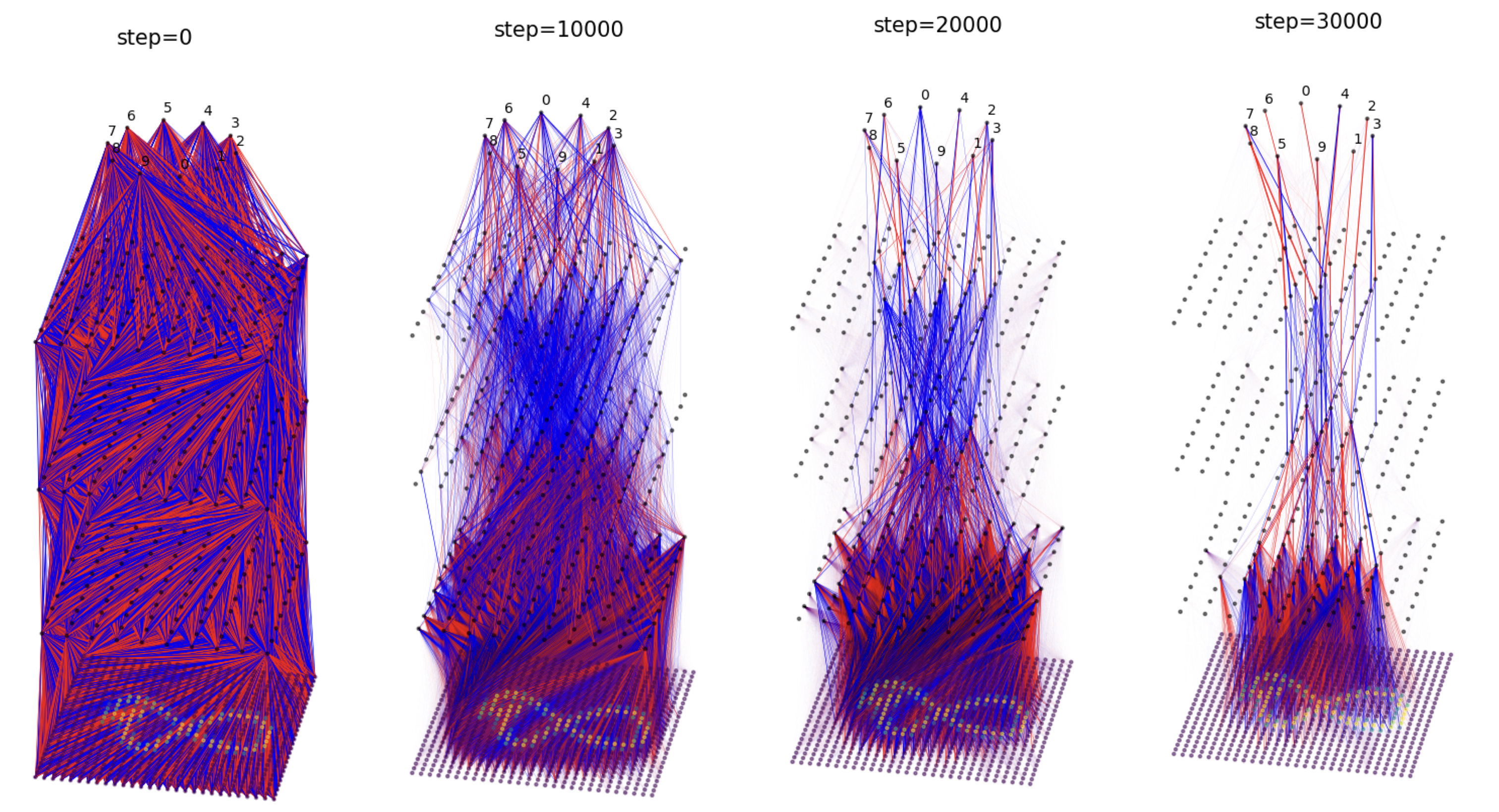}
    \caption{Evolution of a 4 Layer MLP trained with BIMT.}
    \label{fig:mnist_4L}
\end{figure}

{\bf MLPs with other depths} In the main text, we showed the results for a 3 Layer MLP. We also show the results (how the connectivity graphs evolve in training) for a 2 Layer MLP and a 4 Layer MLP in Figure~\ref{fig:mnist_2L} and~\ref{fig:mnist_4L}, respectively.

\end{appendix}

\end{document}